\documentclass[conference]{IEEEtran}
\IEEEoverridecommandlockouts

\usepackage{times}

\usepackage[numbers]{natbib}
\usepackage{multicol}
\usepackage[bookmarks=true]{hyperref}
\usepackage{amsmath}
\usepackage{graphicx}
\usepackage{caption}
\usepackage{capt-of}
\let\labelindent\relax
\usepackage{enumitem}

\usepackage[T1]{fontenc}

\usepackage[capitalize]{cleveref}
\crefname{figure}{Fig.}{Figs.}
\crefname{section}{Sec.}{Secs.}
\crefname{table}{Tab.}{Tabs.}

\usepackage{listings}
\usepackage{booktabs}
\usepackage{subcaption}

\usepackage{algorithm}
\usepackage{algpseudocode}

\usepackage{xcolor}
\usepackage[colorinlistoftodos,prependcaption,textsize=small]{todonotes}

\begin{document}

\title{Hierarchical Vision-Language Planning for Multi-Step Humanoid Manipulation}

\author{
    André Schakkal$^{1,2}$\thanks{Contact: \texttt{\{schakkal,bzando,ztyang,azizan\}@mit.edu}} \quad 
    Ben Zandonati$^{1}$\quad 
    Zhutian Yang$^{1}$\quad 
    Navid Azizan$^{1}$ \\[1em]
    $^{1}$Massachusetts Institute of Technology\\
    $^{2}$École Polytechnique Fédérale de Lausanne
}

\setcounter{figure}{1}
\makeatletter
\let\@oldmaketitle\@maketitle%
\renewcommand{\@maketitle}{\@oldmaketitle%
  \begin{center}
    \includegraphics[width=\linewidth]{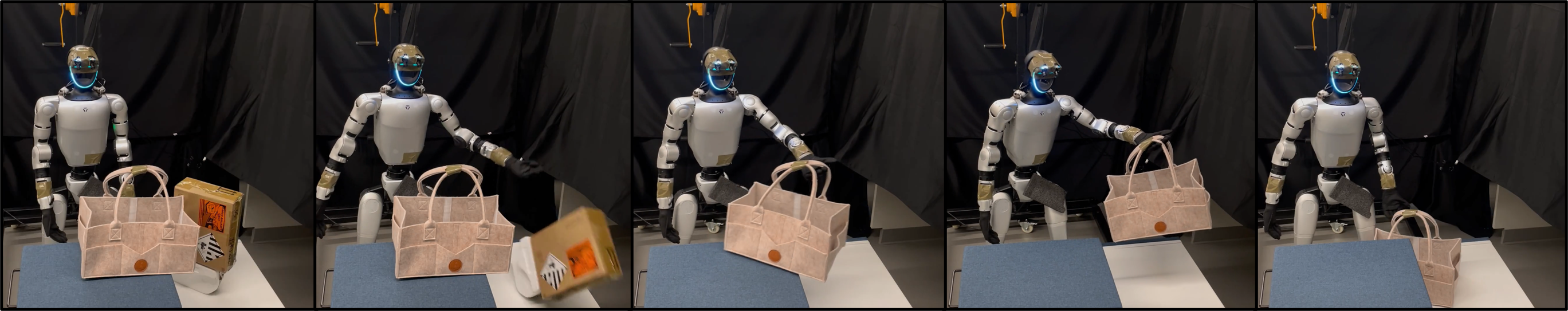} 
    \label{fig:overview}
  \end{center}
  \vspace{-12pt}
  \small{\textbf{Fig.~\thefigure:\label{fig:overview}}~
  Our hierarchical humanoid manipulation system autonomously executes a multi-step rearrangement task. The robot first pushes aside an obstacle, then picks up a bag and places it onto another surface. The system uses pretrained VLMs to orchestrate manipulation policies that output whole-body joint positions, which serve as tracking targets for a low-level RL policy.}

  \medskip}%
\makeatother

\maketitle

\begin{abstract}
Enabling humanoid robots to reliably execute complex multi-step manipulation tasks is crucial for their effective deployment in industrial and household environments. This paper presents a hierarchical planning and control framework designed to achieve reliable multi-step humanoid manipulation.
The proposed system comprises three layers: (1) a low-level controller trained via reinforcement learning responsible for tracking whole-body motion targets,
(2) a mid-level set of skill policies trained via imitation learning that produce motion targets for different steps of a task, 
and (3) a high-level vision-language planning module that determines which skills should be executed and also monitors their completion in real time using pretrained vision-language models (VLMs). 
Experimental validation is performed on a Unitree G1 humanoid robot executing a non-prehensile pick-and-place task. Over 40 real-world trials, the hierarchical system achieved a 73\% success rate in completing the full manipulation sequence. 
These experiments confirm the feasibility of the proposed hierarchical system, highlighting the benefits of VLM-based skill planning and monitoring for multi-step manipulation scenarios.
See \url{https://vlp-humanoid.github.io/} for a video demonstration of the policy rollout.

\end{abstract}

\IEEEpeerreviewmaketitle

\section{Introduction}

Humanoid robots offer a promising path towards general-purpose autonomy, integrated seamlessly into human environments. The humanoid morphology, while advantageous for tool use and ergonomic compatibility, presents challenging control and planning problems \cite{gu_humanoid_2025}. Recent reinforcement-learning (RL) and imitation-learning (IL) breakthroughs have produced striking single-skill demonstrations such as 
walking, dancing, or boxing \cite{cheng_expressive_2024,fu_humanplus_2024,he_learning_2024-1,he_omnih2o_2024,he_asap_2025}. Yet, practical applications require that the humanoid carry out multi-step tasks, i.e., interacting with multiple objects and regions over time. 
The sequencing and integration of these learned skills remains challenging, due to the high-dimensional nature of humanoid control, the requirement for continuous real-time visual feedback, and the complexity of determining precise transitions between consecutive skills.

To bridge the gap between isolated skill execution and long-horizon autonomy, robots must combine low-level motor dexterity with higher-level reasoning and task monitoring. 
Vision-language models
(VLMs) present a unique opportunity to integrate planning and monitoring with robust learned control policies, enabling embodied humanoid agents to execute extended, multi-step manipulation tasks in unstructured, human environments \cite{dalal_local_2025, yang2024guiding, team_gemini_2024}.

In this paper, we present a hierarchical framework for humanoid manipulation planning and control, spanning three layers. 
The lowest level employs a whole-body tracking controller trained via proximal policy optimization (PPO) \cite{schulman_proximal_2017}, which tracks motion targets provided by a mid-level imitation learning (IL) policy. This mid-level policy is trained on human teleoperation data, where human poses are retargeted onto the robot's morphology. Each IL-based skill policy maps egocentric visual observations and proprioceptive states to full-body motion commands, enabling the execution of complex loco-manipulation tasks. 
At the highest level, we propose a hierarchical planning module consisting of two VLMs: a low-frequency VLM planner is used to generate sequences of IL skills from visual and textual task inputs, and a higher-frequency 
VLM-based skill-execution monitor that continuously verifies skill completion, orchestrating transitions between skills. 

We empirically validate our proposed hierarchical system on the 29-DoF Unitree G1 humanoid robot, performing a representative multi-step pick-and-place task. In real-world testing, the integrated system successfully completed the full manipulation sequence in 73\% of 40 trials. Our hierarchical approach demonstrates effective coordination of manipulation skills and visual feedback, highlighting the benefit of structured skill decomposition and VLM-based execution monitoring for reliable, multi-step humanoid autonomy.

The primary contributions of our work are: (1) a novel hierarchical vision-language planning and monitoring framework that dynamically sequences and verifies humanoid manipulation skills; and (2) an integrated autonomous system validated on a real humanoid robot performing complex multi-step tasks in realistic environments, demonstrating practical feasibility and effectiveness.

\section{Related Work}

\subsection{Hierarchical Control in Humanoids}

Recent approaches to humanoid control commonly split the control task into two separate layers:  a low-level reinforcement learning (RL) controller that tracks reference motion targets, and a higher-level module that generates these motion targets.
ExBody \cite{cheng_expressive_2024} first showed that a PPO tracker trained in Isaac Gym \cite{makoviychuk_isaac_2021} can reproduce upper-body MoCap clips on the Unitree H1 while leaving the legs free for balance. ExBody2 \cite{ji_exbody2_2025} extends the idea to whole-body loco-manipulation via teacher-student distillation and domain randomization, achieving high-fidelity whole-body tracking on both H1 and G1 platforms. HumanPlus \cite{fu_humanplus_2024} replaces the multilayer perceptron (MLP) backbone of the low-level policy with a transformer backbone.  Additionally, it expands the training data to use the AMASS dataset \cite{mahmood_amass_2019}, significantly increasing the diversity of available human motion examples.

While low-level policies reliably track motion targets, generating these references is equally critical for accomplishing manipulation tasks. H2O (Human-to-Humanoid) \cite{he_learning_2024} utilized single-camera human pose estimation to enable zero-shot teleoperation. OmniH2O \cite{he_omnih2o_2024} extended H2O’s concept, offering a unified interface for teleoperation using various modalities, including RGB cameras, VR, motion capture suits, and exoskeleton systems. OmniH2O further incorporated diffusion-based imitation learning \cite{chi_diffusion_2024}, enabling the autonomous acquisition of manipulation skills directly from human demonstrations. Similarly, HumanPlus \cite{fu_humanplus_2024} adopted this hierarchical structure, leveraging human-teleoperated demonstrations recorded along with proprioceptive and egocentric visual data. These demonstrations were used to train the Humanoid Imitation Transformer (HIT)—a transformer-based imitation learning policy inspired by the Action Chunking Transformer (ACT) \cite{zhao_learning_2023}. Using only 20–40 demonstrations per skill, HumanPlus reported 60–100\% success rates across diverse whole-body manipulation tasks.

Across these studies, a common hierarchical structure appears, featuring a robust RL-based low-level policy for tracking motion references and a high-level module generating these references through teleoperation or imitation learning. Although this approach has shown success in single-skill tasks (e.g., dancing, lifting, and boxing), existing systems generally require human intervention when transitioning between skills, thus limiting their ability to operate autonomously over extended task sequences.

\subsection{VLMs for Long-Horizon Manipulation}

Executing multi-step tasks demands semantic understanding, robust memory, and resilience to errors. Recent approaches leveraging large vision-language-action (VLA) models, such as RT-2 \cite{brohan_rt-2_2023}, OpenVLA \cite{kim_openvla_2024}, and $\pi_{0.5}$ \cite{black_pi05_2025}, demonstrate promise in end-to-end planning but typically require large-scale paired datasets and lack interpretability. To address these limitations, hierarchical approaches integrate intermediate semantic representations, bridging abstract instructions and concrete actions (e.g., RT-H \cite{belkhale_rt-h_2024}, NaVILA \cite{cheng_navila_2025}). Another direction employs explicit hierarchies, where high-level vision-language models predict coarse geometric plans refined by lower-level controllers, effectively separating semantic reasoning from robot dynamics \cite{li_hamster_2025,shi_hi_2025}. Additionally, methods like KALM \cite{fang_keypoint_2024}, RoboPoint \cite{yuan_robopoint_2024}, and ManipGen \cite{dalal_local_2025} use VLMs to predict spatial affordances or parameterized skill libraries, facilitating efficient zero-shot generalization.

Despite the advances in humanoid control, current humanoid systems lack integrated high-level modules capable of autonomously selecting and sequencing skills, as well as verifying successful execution. Addressing this gap, our work proposes an additional third hierarchical layer, combining a VLM-based planner and skill monitor on top of the established two-layer humanoid control stacks, enabling fully autonomous execution of extended loco-manipulation tasks.

 \section{Method}

\begin{figure*}[t]
    \centering
    \includegraphics[width=\textwidth]{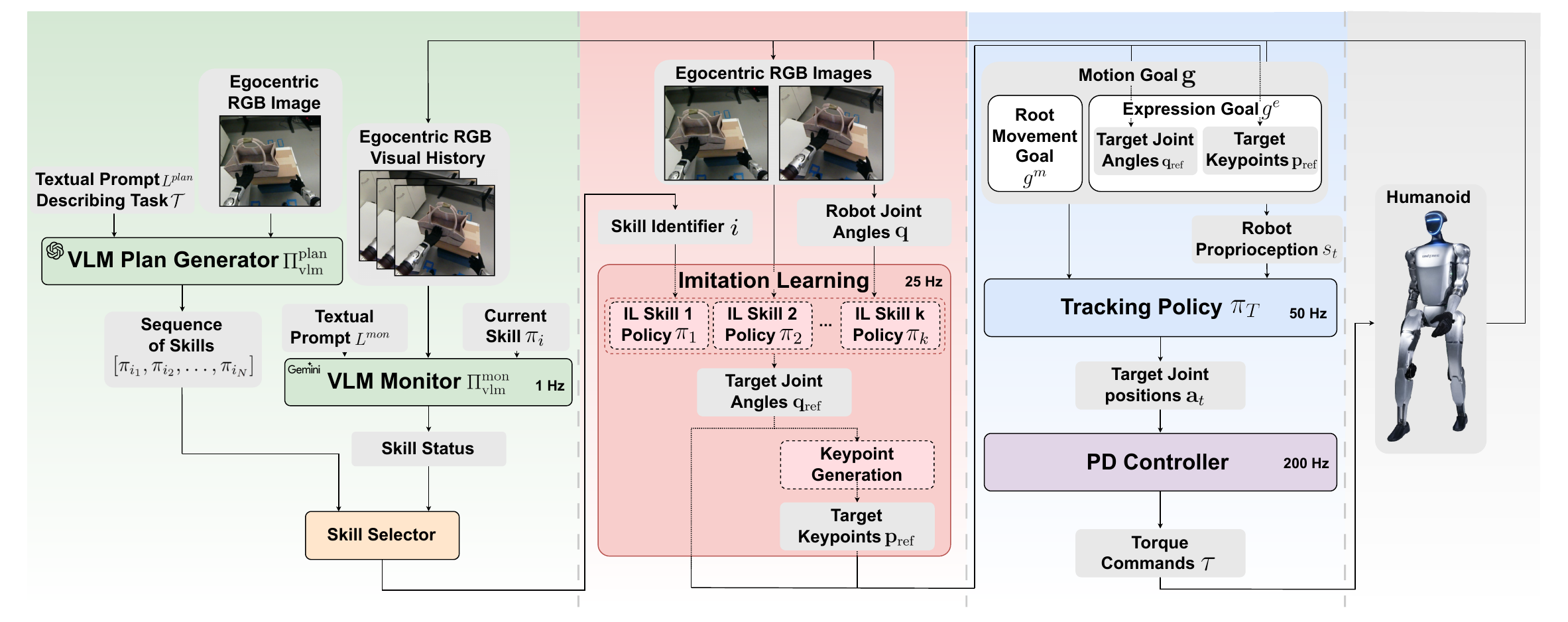}
    \caption{Overview of the proposed hierarchical framework for autonomous multi-step humanoid manipulation. The system consists of: (1) a high-level planning module using vision-language models (VLMs) to generate task-specific skill sequences and monitor their execution via visual feedback, (2) a mid-level motion-generation module composed of imitation-learned action expert policies, mapping egocentric visual inputs and proprioceptive robot states into target joint configurations, and (3) a low-level reinforcement-learning-based tracking policy generating feasible joint-angle trajectories, executed by a high-frequency PD controller.}
    \label{fig:full_method_diagram}
\end{figure*}

We propose a three-layer hierarchical control and planning framework for multi-step humanoid manipulation, illustrated in Fig.~\ref{fig:full_method_diagram}. The system consists of: (1) a low-level whole-body tracking controller trained via reinforcement learning (Sec.~\ref{sec:low-level}), (2) a mid-level motion-generation module composed of multiple imitation-learned action expert policies, each trained from teleoperated demonstrations (Sec.~\ref{sec:mid-level}), and (3) a high-level vision-language-based planning and execution monitoring module (Sec.~\ref{sec:high-level}). The following sections detail each component.

\subsection{Low-Level Tracking Policy}\label{sec:low-level}

Our low-level module is a reinforcement learning-based tracking controller designed to track generated motion goals. Building upon the ExBody architecture~\cite{cheng_expressive_2024}, we extended the implementation to the full 29-DoF Unitree G1 humanoid robot, incorporating whole-body tracking.

\subsubsection{Problem Formulation}

The tracking policy, denoted as
\[
\pi_T: G \times S \rightarrow A,
\]
maps a motion goal \(G\) and the robot's current state \(S\) to feasible joint commands \(A\). The goal space \(G\) consists of two components: a Root Motion Goal \(G_m\) defined by a desired base linear velocity \(v_{\text{ref}}\) and orientation \(\omega_{\text{ref}}\), and an Expression Goal \(G_e\) defined by target joint angles \(q_{\text{ref}}\) and a set of 3D keypoint positions \(p_{\text{ref}}\) corresponding to specific body landmarks, the expression goal specifies the desired motion or gesture that the robot should track. The state \(S\) includes proprioceptive observations such as joint positions and velocities, IMU data, and a short history of previous states. The policy outputs target joint angles at \(50\,\text{Hz}\), enforced using a high-frequency PD controller operating at \(200\,\text{Hz}\).

\subsubsection{Policy Architecture and Training}

We employ a standard actor-critic architecture trained via proximal policy optimization (PPO)~\cite{schulman_proximal_2017}. Both the actor and critic networks are implemented as three-layer MLPs (256 hidden units, ReLU activations). Training occurs within the Isaac Gym simulator using 4096 parallel environments. To ensure robustness in real-world conditions, we extensively randomize environment parameters, including gravity, friction, mass distribution, motor strength, external perturbations, and terrain irregularities.

The reward function combines tracking accuracy terms, adapted from ExBody~\cite{cheng_expressive_2024} and ExBody2~\cite{ji_exbody2_2025}, along with additional stability-promoting regularization terms specifically tuned for the G1 robot. Detailed reward components are provided in \cref{app:low-level}.

\subsubsection{Training Data}

Training data was sourced from a curated subset of the AMASS dataset \cite{mahmood_amass_2019}, selected to cover the robot's operational range. The data was supplemented by additional teleoperation demonstrations that we collected, tailored to the specific downstream skills and tasks outlined in the project. This combination ensures both diversity and task-relevance of the training data. 
Quantitative and qualitative evaluations of the tracking performance of the low-level policy outlined in \cref{app:low-level}.

\subsection{Mid-Level Imitation Learned Skills} \label{sec:mid-level}

Building upon the low-level tracking policy's capability to execute motion goals precisely, the mid-level module generates higher-level skill behaviors autonomously. Specifically, it translates sensory inputs into coherent joint-motion targets that serve as reference trajectories for the tracking policy. This module integrates two complementary capabilities: an RGB-based teleoperation pipeline for demonstration collection, and imitation-learned (IL) expert policies trained from these demonstrations to execute individual skills autonomously.

\subsubsection{Teleoperation via Human Pose Retargeting}

To facilitate the collection of demonstration data, we developed an RGB-based teleoperation pipeline allowing human operators to control the humanoid robot directly. This system consists of three primary stages: human pose estimation from RGB images, retargeting of human poses onto the robot, and generation of robot-specific keypoints.

For human pose estimation, we employ HybrIK (Hybrid Inverse Kinematics) \cite{li_hybrik_2022}, a hybrid analytical-neural inverse kinematics model. HybrIK initially predicts 3D joint positions from single-view RGB inputs using a convolutional neural network (CNN). Subsequently, these predictions are refined through a physics-informed inverse kinematics optimization step, yielding temporally smooth and anatomically plausible human joint configurations. This pipeline eliminates the need for specialized motion-capture equipment, requiring only a standard RGB camera setup.

Due to morphological differences between human operators and the Unitree G1 robot—including limb proportions, joint limits, and degrees of freedom—a dedicated retargeting procedure translates the human pose estimates into robot-compatible joint configurations. 
This retargeting procedure is outlined in \cref{app:low-level}.

As a final step, the low-level tracking policy expects as input both joint-angle references and corresponding 3D keypoints representing the center-of-mass (CoM) positions of selected robot body links. To generate these keypoints, we apply forward kinematics in simulation: we first set the robot to the retargeted joint configuration, and then compute the spatial coordinates of the predefined body positions. The resulting joint angles and keypoints form the complete motion reference that can be directly executed by the tracking controller.

\subsubsection{Imitation Learning Module}

Building upon the established teleoperation pipeline, we next develop the imitation learning (IL) module, the mid-level component responsible for generating autonomous motion targets from onboard sensory inputs.

While the low-level tracking policy executes externally provided motion references, the IL module produces these references directly from sensory inputs, enabling robot autonomy. Formally, the IL policy is defined as a function
\[
\pi_{IL}: \mathcal{X}\rightarrow\mathcal{A},
\]
where the observation state at each timestep $t$ is
\[
x_t = (\mathbf{q}_t,\,I_t^{(1)},\,I_t^{(2)}),
\]
comprising proprioceptive joint angles $\mathbf{q}_t$ and binocular egocentric RGB images $I_t^{(1)} $ and $I_t^{(2)}$. The action space $\mathcal{A}$ consists of future joint configurations $\mathbf{q}$, directly interfacing with the low-level tracking policy. 

For the IL policies, we adopt the Humanoid Imitation Transformer (HIT)~\cite{fu_humanplus_2024}, designed specifically for high-DoF humanoids with binocular vision. It is based on the Action Chunking Transformer (ACT)~\cite{zhao_learning_2023} model, which has demonstrated improved robotic control by predicting multi-step future actions in one forward pass. 
At inference, HIT outputs action chunks of 50 joint-angle targets at 25~Hz. During training, an auxiliary $L_2$ loss between predicted and actual future visual embeddings improves visual grounding and generalization.
The predicted joint angles are subsequently transformed into corresponding 3D keypoints via forward kinematics, matching the inputs required by the low-level tracking controller.

Instead of using the robot's onboard Intel RealSense camera, we employ two externally mounted ELP high-speed RGB cameras with wide-angle lenses. This binocular configuration provides consistent frame rates and explicit depth information, enhancing spatial reasoning capabilities essential for manipulation tasks. 

Demonstration data for IL training is collected through teleoperation, forming a dataset
\[
\mathcal{D} = \{(x_1,a_1), (x_2,a_2),\dots,(x_N,a_N)\},
\]
containing synchronized state-action pairs recorded at 25~Hz. We collect data for the representative task rearrangement involving 
picking a bag from a table and placing it onto another. Only successful sequences (30 in total) are retained, segmented into separate skill-specific datasets, i.e., picking and placing, for training two policies. IL policy training employs the demonstration data with hyperparameters specified in Table~\ref{tab:hit_training_params}. After training, we evaluated independently each skill policy over 30 autonomous trials, which showed success rates of 
$90\%$ for picking and $83\%$ for placing.

\subsection{Multi-Step Skill Planning and Execution Monitoring}
\label{sec:high-level}

While imitation-learned skills enable reliable execution of short-term actions, real-world humanoid tasks frequently require combining multiple skills into longer sequences to achieve complex goals. We address this by introducing a high-level planning and execution monitoring module, capable of dynamically selecting and verifying the execution of skill sequences. Our approach leverages pretrained vision-language models (VLMs) embedded within a closed-loop \emph{planner-monitor} architecture.

To clearly distinguish conceptual terminology, we define:
\begin{itemize}
    \item A \textbf{skill} as an individual, short-duration manipulation capability (e.g., picking or placing an object).
    \item A \textbf{task} as a higher-level objective involving at least two skills executed sequentially that change the state of objects in the world.
\end{itemize}

Our hierarchical system thus plans at the task-level while monitoring and executing at the skill-level.

\subsubsection{Problem Formulation}

Formally, a task is defined by a natural-language goal $g$ (e.g., ``Pick up the bag and place it on the table'') and an initial visual observation $o_0$. The goal is to autonomously generate and reliably execute a sequence of parameterized skills $\sigma = [\pi^{(1)}, \pi^{(2)}, \dots, \pi^{(N)}]$ to satisfy the conditions described by $g$.

\subsubsection{System Architecture}

Our high-level planning and monitoring module consists of two complementary components:

\begin{enumerate}[leftmargin=1.2cm,label=(\alph*)]
    \item \textbf{VLM Planner ($\mathcal{P}$)}: A GPT-4o model~\cite{openai_gpt-4o_2024} generates structured, interpretable skill sequences from visual and textual task inputs.
    \item \textbf{VLM Skill Monitor ($\mathcal{M}$)}: A lightweight Gemini-2.0-Flash-Lite model~\cite{team_gemini_2024} continuously verifies the completion of each executed skill at approximately 1\,Hz.
\end{enumerate}

Together, these two components form an iterative planning-monitoring loop, ensuring coherent multi-step task execution.

\subsubsection{Skill Library and PDDL-like Representation}

We represent each skill available to the robot using structured, human-readable descriptions that include the `preconditions' and `effects' fields of an operator definition in Planning Domain Definition Language (PDDL)~\cite{aeronautiques1998pddl}. Each skill description includes a one-sentence summary, explicit natural-language preconditions, effects, and example verification queries, as illustrated for the \texttt{Pick} skill in Fig.~\ref{fig:description}. Unlike rigid symbolic representations, this PDDL-like form balances logical rigor and flexibility, enabling the planner to perform common-sense reasoning without exhaustive state enumeration.

\begin{figure}[H]
    \centering
    \begin{lstlisting}
Pick:
Description: 'Pick up an object and hold it up'
Preconditions: [
    'hand is empty', 
    'object is graspable', 
    'object is on a surface'
]
Effects: [
    'robot is holding object up', 
    'object is no longer on the surface'
]
Examples: 'Has the robot successfully picked up and is now holding the bag?'
\end{lstlisting}
    \caption{Example structured skill description provided to the VLM Planner and Skill Monitor.}
    \label{fig:description}
\end{figure}

\subsubsection{VLM Planner}

Given an initial image $o_0$, task instruction $g$, and the structured skill library, the VLM Planner $\mathcal{P}$ outputs a grounded sequence of executable skills $\sigma$. GPT-4o generates the skill sequence by visually grounding skill preconditions and effects, translating them into binary visual-question-answering queries for logical coherence. For instance, the planner implicitly understands causal ordering, such as the necessity of picking up an object before placing it elsewhere, from the provided natural-language annotations, without explicit symbolic planning constraints.

\subsubsection{VLM Skill Monitor}

The VLM Skill Monitor $\mathcal{M}$ continuously evaluates the execution state through real-time visual feedback. Operating at approximately 1\,Hz, it assesses short video snippets (10–15 frames extracted from 1.5-second segments) captured during skill execution. For each skill, the monitor uses the provided natural-language verification queries to output a binary decision: either \emph{completed} or \emph{in progress}. This approach ensures timely and reliable skill transitions while maintaining real-time responsiveness.

A walkthrough example of the functioning of this module is presented in Appendix \ref{app:example} while Appendix \ref{app:prompts} shows example VLM prompts.

\subsection{Full Hierarchical System}

The complete integrated system implements the hierarchical pipeline presented in Fig.~\ref{fig:full_method_diagram}. The pipeline comprises: (1) a \textbf{high-level VLM planner} (GPT-4o) that generates a sequence of skills from initial visual inputs and natural-language task instructions, (2) a real-time \textbf{VLM skill monitor} (Gemini-2.0-Flash-Lite) that verifies skill execution status based on visual feedback, (3) a \textbf{skill selector} that dynamically chooses the current skill to execute based on the planned sequence and monitor feedback, (4) a \textbf{mid-level imitation learning Module} that activates the corresponding imitation-learned policy according to the selected skill, producing joint-motion targets from egocentric images and proprioceptive data, and (5) a \textbf{low-level tracking policy} combined with a high-frequency \textbf{PD controller}, converting these joint targets into executable motor torque commands.

A key advantage of this hierarchical design is its inherent modularity: individual IL skills are trained independently and added directly to the skill library, enabling effortless expansion of the robot’s behavioral repertoire. Newly acquired skills become immediately available for integration by the high-level planning module, facilitating seamless composition into increasingly complex, multi-step tasks without extensive system reconfiguration.

\section{Experiments}

We conducted experiments to evaluate the effectiveness and robustness of our proposed hierarchical control framework for multi-step humanoid manipulation tasks on a real-world robot platform.

\subsection{Experimental Setup}

Experiments were conducted using the 29-degree-of-freedom (DoF) Unitree G1 humanoid robot. The robot was equipped with two externally mounted ELP RGB cameras, providing binocular egocentric visual inputs for the mid-level imitation learning (IL) policies. The robot was tested within a controlled environment consisting of two tables and objects arranged to simulate representative manipulation tasks common in household scenarios. 

\subsection{Evaluation Task}

To validate the hierarchical framework, we defined a representative multi-step manipulation task involving:

\begin{enumerate}[leftmargin=*]
    \item \textbf{Picking}: The robot must grasp and lift the bag from the initial table.
    \item \textbf{Placing}: The robot needs to accurately place the bag onto a second table.
\end{enumerate}

\subsection{Experimental Procedure}

A total of 40 independent trials were conducted. The robot operated fully autonomously, executing plans generated by the high-level VLM planner, with skills dynamically selected and monitored for completion in real-time by the VLM execution monitor. Success was defined by the robot correctly picking up the bag and successfully placing it on the designated target surface.

\subsection{Experimental Results}

\begin{table}[]\centering
\caption{Evaluation of individual policies and the full system.}
\begin{tabular}{@{}llll@{}}
\toprule
                    & Pick & Place  & Pick-Place \\ \midrule
Number of trials    & 30   & 30     & 40         \\
Number of successes & 27   & 25     & 29         \\
Success rate        & 90\% & 83\% & 73\%     \\ \bottomrule
\end{tabular}
\label{tab:results}
\end{table}

Over 40 trials, the integrated hierarchical system successfully completed the entire manipulation sequence with an overall success rate of \(73\%\), as summarized in Table~\ref{tab:results}. 
During each trial, we recorded failures at three distinct levels of the hierarchical framework. The observed failures were categorized as follows, in order of frequency:

\begin{itemize}[leftmargin=*]
    \item \textbf{Skill Policy Failure}: Occurred when a mid-level imitation policy failed to execute the desired skill. Most frequently occurred during the grasping phase, primarily due to variations in object positioning beyond training data distributions.
    \item \textbf{VLM Execution Monitor Failure}: Instances where the visual verification prematurely indicated task completion due to subtle positioning inaccuracies or ambiguous visual cues.
    \item \textbf{VLM Planner Failure}: Occasional mis-grounding issues occurred, resulting in the generation of incorrect skill sequences.
\end{itemize}

Qualitatively, the robot exhibited stable and coherent manipulation behaviors across successful trials. Skills transitioned smoothly under guidance from the VLM execution monitor, demonstrating effective real-time adaptation and corrective feedback. Minor irregularities in the skills arose due to executing multiple consecutive 50-step action chunks, causing subtle positional resets between chunks. This suggests that introducing a smoothing or blending function between consecutive chunks could further enhance motion fluidity.

\section{Discussion}

The experiments demonstrate that augmenting the canonical two-layer humanoid stack with a vision-language planning and monitoring layer enables reliable execution of multi-step non-prehensile manipulation routines in the physical world. Deployed on a 29-DoF Unitree G1, our proposed three-level architecture successfully completed the obstructed bag-transfer task in 73\% of forty independent trials. The execution monitor's one-second mean verification latency at 1Hz fits comfortably within the robot's motion cycle, and coexists with high-frequency skill policies and RL controller, illustrating that the large-model reasoning and perception need not preclude high-rate control.

Compared to prior humanoid manipulation systems, which often rely on open-loop sequences of predefined skills, our approach introduces continuous closed-loop reasoning. By integrating interpretable VLM-based planning with real-time visual verification, our framework dynamically selects appropriate skills and explicitly determines when transitions between skills should occur.

Furthermore, unlike purely end-to-end vision-language-action (VLA) approaches, which depend heavily on large-scale, paired demonstration datasets and typically lack interpretability, our explicit planner-monitor design provides clear introspection into decision-making processes. This transparency facilitates targeted debugging, allowing developers to easily diagnose and address failures. Similarly, compared to strictly symbolic Task and Motion Planning (TAMP) frameworks, our system’s reliance on soft, natural-language-based skill descriptions—akin to a flexible version of PDDL—greatly simplifies the integration and extension of new behaviors without necessitating detailed domain-specific expertise.

\subsection{Failures and Limitations}

A qualitative failure analysis reveals three dominant modes. First, skill-policy drift occurs when partial occlusions or out-of-distribution object poses cause the imitation policy to stall short of a grasp or placement. Second, planner misgrounding emerges when the deliberative VLM hallucinates objects or fails to sequence skills correctly. This typically involves adding additional irrelevant plan steps or formulating redundant transition questions. Third, optimistic monitor judgments occasionally trigger a premature transition when subtle geometric conditions---such as a bag resting on the edge of a table---are misclassified as success.

Several limitations temper the present findings. Our chosen evaluation task family, while non-trivial, is confined to table-top pick-and-place scenes and therefore under-samples the combinatorial diversity of domestic manipulation. The 1Hz verification rate, although adequate for deliberative motions, would be insufficient for fast skills, such as rapid recovery motions. This implicitly defines the timescale of skill abstraction. 

\subsection{Future Work} 

Our modular architecture provides clear pathways for future enhancements:

\textbf{Error detection and adaptive re-planning:}~~A critical future enhancement includes adapting the VLM execution monitor system to explicitly detect execution failures or anomalous states \cite{park2024quantifying,sharma2021sketching}. Through failure detection, we can trigger real-time adaptive replanning. This promises to improve the robustness of the system.

\textbf{Enhanced semantic and contextual conditioning:}~~Future work involves incorporating richer conditioning beyond visual and proprioceptive states from the high-level execution monitor. For example, replacing the ResNet backbone with that of CLIP \cite{radford_learning_2021} would allow the VLM to have finer-grained semantic control of the skills via text-conditioning. 

\textbf{Expanding the skill library:}~~Finally, the current skill library is focused primarily on pick-and-place. An immediate next step involves expanding the robot's skill repertoire. Examples include non-prehensile bi-manipulation and tool use. This broader skill set would enable more complex, realistic long-horizon tasks, requiring the planning and monitoring components to handle increasingly diverse and dynamic environments.

\section{Conclusion}

In this work, we presented a hierarchical planning and control framework for multi-step humanoid manipulation tasks, combining robust RL-based whole-body tracking, imitation-learned skill policies, and a high-level vision-language planning and monitoring module. Evaluated on the Unitree G1 humanoid robot, our integrated system demonstrated successful autonomous execution of complex manipulation sequences, achieving a 73\% success rate across real-world trials. Crucially, our approach offers inherent modularity, interpretability, and adaptability---enabling straightforward extension of the robot’s skill repertoire. This work highlights the effectiveness of integrating vision-language models with hierarchical control, paving the way toward increasingly capable autonomous humanoid systems operating in dynamic human environments.

\bibliographystyle{plainnat}
\bibliography{references}

\begin{thebibliography}{31}
\providecommand{\natexlab}[1]{#1}
\providecommand{\url}[1]{\texttt{#1}}
\expandafter\ifx\csname urlstyle\endcsname\relax
  \providecommand{\doi}[1]{doi: #1}\else
  \providecommand{\doi}{doi: \begingroup \urlstyle{rm}\Url}\fi

\bibitem[Aeronautiques et~al.(1998)Aeronautiques, Howe, Knoblock, McDermott, Ram, Veloso, Weld, Sri, Barrett, Christianson, et~al.]{aeronautiques1998pddl}
Constructions Aeronautiques, Adele Howe, Craig Knoblock, ISI~Drew McDermott, Ashwin Ram, Manuela Veloso, Daniel Weld, David~Wilkins Sri, Anthony Barrett, Dave Christianson, et~al.
\newblock Pddl| the planning domain definition language.
\newblock \emph{Technical Report, Tech. Rep.}, 1998.

\bibitem[Belkhale et~al.(2024)Belkhale, Ding, Xiao, Sermanet, Vuong, Tompson, Chebotar, Dwibedi, and Sadigh]{belkhale_rt-h_2024}
Suneel Belkhale, Tianli Ding, Ted Xiao, Pierre Sermanet, Quon Vuong, Jonathan Tompson, Yevgen Chebotar, Debidatta Dwibedi, and Dorsa Sadigh.
\newblock {RT}-{H}: {Action} {Hierarchies} {Using} {Language}.
\newblock arXiv, June 2024.
\newblock \doi{10.48550/arXiv.2403.01823}.
\newblock arXiv:2403.01823 [cs].

\bibitem[Black et~al.(2025)Black, Brown, Darpinian, Dhabalia, Driess, Esmail, Equi, Finn, Fusai, Galliker, Ghosh, Groom, Hausman, Ichter, Jakubczak, Jones, Ke, LeBlanc, Levine, Li-Bell, Mothukuri, Nair, Pertsch, Ren, Shi, Smith, Springenberg, Stachowicz, Tanner, Vuong, Walke, Walling, Wang, Yu, and Zhilinsky]{black_pi05_2025}
Kevin Black, Noah Brown, James Darpinian, Karan Dhabalia, Danny Driess, Adnan Esmail, Michael Equi, Chelsea Finn, Niccolo Fusai, Manuel~Y. Galliker, Dibya Ghosh, Lachy Groom, Karol Hausman, Brian Ichter, Szymon Jakubczak, Tim Jones, Liyiming Ke, Devin LeBlanc, Sergey Levine, Adrian Li-Bell, Mohith Mothukuri, Suraj Nair, Karl Pertsch, Allen~Z. Ren, Lucy~Xiaoyang Shi, Laura Smith, Jost~Tobias Springenberg, Kyle Stachowicz, James Tanner, Quan Vuong, Homer Walke, Anna Walling, Haohuan Wang, Lili Yu, and Ury Zhilinsky.
\newblock Pi0.5: a {Vision}-{Language}-{Action} {Model} with {Open}-{World} {Generalization}.
\newblock arXiv, April 2025.
\newblock \doi{10.48550/arXiv.2504.16054}.
\newblock arXiv:2504.16054 [cs].

\bibitem[Brohan et~al.(2023)Brohan, Brown, Carbajal, Chebotar, Chen, Choromanski, Ding, Driess, Dubey, Finn, Florence, Fu, Arenas, Gopalakrishnan, Han, Hausman, Herzog, Hsu, Ichter, Irpan, Joshi, Julian, Kalashnikov, Kuang, Leal, Lee, Lee, Levine, Lu, Michalewski, Mordatch, Pertsch, Rao, Reymann, Ryoo, Salazar, Sanketi, Sermanet, Singh, Singh, Soricut, Tran, Vanhoucke, Vuong, Wahid, Welker, Wohlhart, Wu, Xia, Xiao, Xu, Xu, Yu, and Zitkovich]{brohan_rt-2_2023}
Anthony Brohan, Noah Brown, Justice Carbajal, Yevgen Chebotar, Xi~Chen, Krzysztof Choromanski, Tianli Ding, Danny Driess, Avinava Dubey, Chelsea Finn, Pete Florence, Chuyuan Fu, Montse~Gonzalez Arenas, Keerthana Gopalakrishnan, Kehang Han, Karol Hausman, Alexander Herzog, Jasmine Hsu, Brian Ichter, Alex Irpan, Nikhil Joshi, Ryan Julian, Dmitry Kalashnikov, Yuheng Kuang, Isabel Leal, Lisa Lee, Tsang-Wei~Edward Lee, Sergey Levine, Yao Lu, Henryk Michalewski, Igor Mordatch, Karl Pertsch, Kanishka Rao, Krista Reymann, Michael Ryoo, Grecia Salazar, Pannag Sanketi, Pierre Sermanet, Jaspiar Singh, Anikait Singh, Radu Soricut, Huong Tran, Vincent Vanhoucke, Quan Vuong, Ayzaan Wahid, Stefan Welker, Paul Wohlhart, Jialin Wu, Fei Xia, Ted Xiao, Peng Xu, Sichun Xu, Tianhe Yu, and Brianna Zitkovich.
\newblock {RT}-2: {Vision}-{Language}-{Action} {Models} {Transfer} {Web} {Knowledge} to {Robotic} {Control}.
\newblock arXiv, July 2023.
\newblock \doi{10.48550/arXiv.2307.15818}.
\newblock arXiv:2307.15818 [cs].

\bibitem[Cheng et~al.(2025)Cheng, Ji, Yang, Gongye, Zou, Kautz, Bıyık, Yin, Liu, and Wang]{cheng_navila_2025}
An-Chieh Cheng, Yandong Ji, Zhaojing Yang, Zaitian Gongye, Xueyan Zou, Jan Kautz, Erdem Bıyık, Hongxu Yin, Sifei Liu, and Xiaolong Wang.
\newblock {NaVILA}: {Legged} {Robot} {Vision}-{Language}-{Action} {Model} for {Navigation}.
\newblock arXiv, February 2025.
\newblock \doi{10.48550/arXiv.2412.04453}.
\newblock arXiv:2412.04453 [cs].

\bibitem[Cheng et~al.(2024)Cheng, Ji, Chen, Yang, Yang, and Wang]{cheng_expressive_2024}
Xuxin Cheng, Yandong Ji, Junming Chen, Ruihan Yang, Ge~Yang, and Xiaolong Wang.
\newblock Expressive {Whole}-{Body} {Control} for {Humanoid} {Robots}.
\newblock arXiv, March 2024.
\newblock \doi{10.48550/arXiv.2402.16796}.
\newblock arXiv:2402.16796 [cs].

\bibitem[Chi et~al.(2023)Chi, Xu, Feng, Cousineau, Du, Burchfiel, Tedrake, and Song]{chi_diffusion_2024}
Cheng Chi, Zhenjia Xu, Siyuan Feng, Eric Cousineau, Yilun Du, Benjamin Burchfiel, Russ Tedrake, and Shuran Song.
\newblock Diffusion policy: Visuomotor policy learning via action diffusion.
\newblock page 02783649241273668. SAGE Publications Sage UK: London, England, 2023.

\bibitem[Dalal et~al.(2025)Dalal, Liu, Talbott, Chen, Pathak, Zhang, and Salakhutdinov]{dalal_local_2025}
Murtaza Dalal, Min Liu, Walter Talbott, Chen Chen, Deepak Pathak, Jian Zhang, and Ruslan Salakhutdinov.
\newblock Local {Policies} {Enable} {Zero}-shot {Long}-horizon {Manipulation}.
\newblock arXiv, March 2025.
\newblock \doi{10.48550/arXiv.2410.22332}.
\newblock arXiv:2410.22332 [cs].

\bibitem[Fang et~al.(2024)Fang, Huang, Mao, Shone, Tenenbaum, Lozano-Pérez, and Kaelbling]{fang_keypoint_2024}
Xiaolin Fang, Bo-Ruei Huang, Jiayuan Mao, Jasmine Shone, Joshua~B. Tenenbaum, Tomás Lozano-Pérez, and Leslie~Pack Kaelbling.
\newblock Keypoint {Abstraction} using {Large} {Models} for {Object}-{Relative} {Imitation} {Learning}.
\newblock arXiv, October 2024.
\newblock \doi{10.48550/arXiv.2410.23254}.
\newblock arXiv:2410.23254 [cs].

\bibitem[Fu et~al.(2024)Fu, Zhao, Wu, Wetzstein, and Finn]{fu_humanplus_2024}
Zipeng Fu, Qingqing Zhao, Qi~Wu, Gordon Wetzstein, and Chelsea Finn.
\newblock {HumanPlus}: {Humanoid} {Shadowing} and {Imitation} from {Humans}.
\newblock arXiv, June 2024.
\newblock \doi{10.48550/arXiv.2406.10454}.
\newblock arXiv:2406.10454 [cs].

\bibitem[Gu et~al.(2025)Gu, Li, Shen, Yu, Xie, McCrory, Cheng, Shamsah, Griffin, Liu, Kheddar, Peng, Zhu, Shi, Nguyen, Cheng, Gao, and Zhao]{gu_humanoid_2025}
Zhaoyuan Gu, Junheng Li, Wenlan Shen, Wenhao Yu, Zhaoming Xie, Stephen McCrory, Xianyi Cheng, Abdulaziz Shamsah, Robert Griffin, C.~Karen Liu, Abderrahmane Kheddar, Xue~Bin Peng, Yuke Zhu, Guanya Shi, Quan Nguyen, Gordon Cheng, Huijun Gao, and Ye~Zhao.
\newblock Humanoid {Locomotion} and {Manipulation}: {Current} {Progress} and {Challenges} in {Control}, {Planning}, and {Learning}.
\newblock arXiv, April 2025.
\newblock \doi{10.48550/arXiv.2501.02116}.
\newblock arXiv:2501.02116 [cs].

\bibitem[He et~al.(2024{\natexlab{a}})He, Luo, He, Xiao, Zhang, Zhang, Kitani, Liu, and Shi]{he_omnih2o_2024}
Tairan He, Zhengyi Luo, Xialin He, Wenli Xiao, Chong Zhang, Weinan Zhang, Kris Kitani, Changliu Liu, and Guanya Shi.
\newblock {OmniH2O}: {Universal} and {Dexterous} {Human}-to-{Humanoid} {Whole}-{Body} {Teleoperation} and {Learning}.
\newblock arXiv, June 2024{\natexlab{a}}.
\newblock \doi{10.48550/arXiv.2406.08858}.
\newblock arXiv:2406.08858 [cs].

\bibitem[He et~al.(2024{\natexlab{b}})He, Luo, Xiao, Zhang, Kitani, Liu, and Shi]{he_learning_2024-1}
Tairan He, Zhengyi Luo, Wenli Xiao, Chong Zhang, Kris Kitani, Changliu Liu, and Guanya Shi.
\newblock Learning human-to-humanoid real-time whole-body teleoperation.
\newblock In \emph{2024 IEEE/RSJ International Conference on Intelligent Robots and Systems (IROS)}, pages 8944--8951. IEEE, 2024{\natexlab{b}}.

\bibitem[He et~al.(2025)He, Gao, Xiao, Zhang, Wang, Wang, Luo, He, Sobanbab, Pan, Yi, Qu, Kitani, Hodgins, Fan, Zhu, Liu, and Shi]{he_asap_2025}
Tairan He, Jiawei Gao, Wenli Xiao, Yuanhang Zhang, Zi~Wang, Jiashun Wang, Zhengyi Luo, Guanqi He, Nikhil Sobanbab, Chaoyi Pan, Zeji Yi, Guannan Qu, Kris Kitani, Jessica Hodgins, Linxi~"Jim" Fan, Yuke Zhu, Changliu Liu, and Guanya Shi.
\newblock {ASAP}: {Aligning} {Simulation} and {Real}-{World} {Physics} for {Learning} {Agile} {Humanoid} {Whole}-{Body} {Skills}.
\newblock arXiv, April 2025.
\newblock \doi{10.48550/arXiv.2502.01143}.
\newblock arXiv:2502.01143 [cs].

\bibitem[He et~al.(2024{\natexlab{c}})He, Lei, Ze, Sreenath, Li, and Xu]{he_learning_2024}
Zhengmao He, Kun Lei, Yanjie Ze, Koushil Sreenath, Zhongyu Li, and Huazhe Xu.
\newblock Learning visual quadrupedal loco-manipulation from demonstrations.
\newblock In \emph{2024 IEEE/RSJ International Conference on Intelligent Robots and Systems (IROS)}, pages 9102--9109. IEEE, 2024{\natexlab{c}}.

\bibitem[Ji et~al.(2025)Ji, Peng, Liu, Li, Yang, Cheng, and Wang]{ji_exbody2_2025}
Mazeyu Ji, Xuanbin Peng, Fangchen Liu, Jialong Li, Ge~Yang, Xuxin Cheng, and Xiaolong Wang.
\newblock {ExBody2}: {Advanced} {Expressive} {Humanoid} {Whole}-{Body} {Control}.
\newblock arXiv, March 2025.
\newblock \doi{10.48550/arXiv.2412.13196}.
\newblock arXiv:2412.13196 [cs].

\bibitem[Kim et~al.(2024)Kim, Pertsch, Karamcheti, Xiao, Balakrishna, Nair, Rafailov, Foster, Lam, Sanketi, Vuong, Kollar, Burchfiel, Tedrake, Sadigh, Levine, Liang, and Finn]{kim_openvla_2024}
Moo~Jin Kim, Karl Pertsch, Siddharth Karamcheti, Ted Xiao, Ashwin Balakrishna, Suraj Nair, Rafael Rafailov, Ethan Foster, Grace Lam, Pannag Sanketi, Quan Vuong, Thomas Kollar, Benjamin Burchfiel, Russ Tedrake, Dorsa Sadigh, Sergey Levine, Percy Liang, and Chelsea Finn.
\newblock {OpenVLA}: {An} {Open}-{Source} {Vision}-{Language}-{Action} {Model}.
\newblock arXiv, September 2024.
\newblock \doi{10.48550/arXiv.2406.09246}.
\newblock arXiv:2406.09246 [cs].

\bibitem[Li et~al.(2021)Li, Xu, Chen, Bian, Yang, and Lu]{li_hybrik_2022}
Jiefeng Li, Chao Xu, Zhicun Chen, Siyuan Bian, Lixin Yang, and Cewu Lu.
\newblock Hybrik: A hybrid analytical-neural inverse kinematics solution for 3d human pose and shape estimation.
\newblock In \emph{Proceedings of the IEEE/CVF conference on computer vision and pattern recognition}, pages 3383--3393, 2021.

\bibitem[Li et~al.(2025)Li, Deng, Zhang, Jang, Memmel, Yu, Garrett, Ramos, Fox, Li, Gupta, and Goyal]{li_hamster_2025}
Yi~Li, Yuquan Deng, Jesse Zhang, Joel Jang, Marius Memmel, Raymond Yu, Caelan~Reed Garrett, Fabio Ramos, Dieter Fox, Anqi Li, Abhishek Gupta, and Ankit Goyal.
\newblock {HAMSTER}: {Hierarchical} {Action} {Models} {For} {Open}-{World} {Robot} {Manipulation}.
\newblock arXiv, February 2025.
\newblock \doi{10.48550/arXiv.2502.05485}.
\newblock arXiv:2502.05485 [cs].

\bibitem[Mahmood et~al.(2019)Mahmood, Ghorbani, Troje, Pons-Moll, and Black]{mahmood_amass_2019}
Naureen Mahmood, Nima Ghorbani, Nikolaus~F Troje, Gerard Pons-Moll, and Michael~J Black.
\newblock Amass: Archive of motion capture as surface shapes.
\newblock In \emph{Proceedings of the IEEE/CVF international conference on computer vision}, pages 5442--5451, 2019.

\bibitem[Makoviychuk et~al.(2021)Makoviychuk, Wawrzyniak, Guo, Lu, Storey, Macklin, Hoeller, Rudin, Allshire, Handa, and State]{makoviychuk_isaac_2021}
Viktor Makoviychuk, Lukasz Wawrzyniak, Yunrong Guo, Michelle Lu, Kier Storey, Miles Macklin, David Hoeller, Nikita Rudin, Arthur Allshire, Ankur Handa, and Gavriel State.
\newblock Isaac {Gym}: {High} {Performance} {GPU}-{Based} {Physics} {Simulation} {For} {Robot} {Learning}.
\newblock arXiv, August 2021.
\newblock \doi{10.48550/arXiv.2108.10470}.
\newblock arXiv:2108.10470 [cs].

\bibitem[OpenAI et~al.(2024)OpenAI, Hurst, Lerer, Goucher, Perelman, Ramesh, Clark, Ostrow, Welihinda, Hayes, Radford, Mądry, Baker-Whitcomb, Beutel, et~al.]{openai_gpt-4o_2024}
OpenAI, Aaron Hurst, Adam Lerer, Adam~P. Goucher, Adam Perelman, Aditya Ramesh, Aidan Clark, A.~J. Ostrow, Akila Welihinda, Alan Hayes, Alec Radford, Aleksander Mądry, Alex Baker-Whitcomb, Alex Beutel, et~al.
\newblock {GPT}-4o {System} {Card}.
\newblock arXiv, October 2024.
\newblock \doi{10.48550/arXiv.2410.21276}.
\newblock arXiv:2410.21276 [cs].

\bibitem[Park et~al.(2024)Park, Wang, Ardeshir, and Azizan]{park2024quantifying}
Young-Jin Park, Hao Wang, Shervin Ardeshir, and Navid Azizan.
\newblock Quantifying representation reliability in self-supervised learning models.
\newblock In \emph{Uncertainty in Artificial Intelligence}, pages 2835--2860. PMLR, 2024.

\bibitem[Radford et~al.(2021)Radford, Kim, Hallacy, Ramesh, Goh, Agarwal, Sastry, Askell, Mishkin, Clark, et~al.]{radford_learning_2021}
Alec Radford, Jong~Wook Kim, Chris Hallacy, Aditya Ramesh, Gabriel Goh, Sandhini Agarwal, Girish Sastry, Amanda Askell, Pamela Mishkin, Jack Clark, et~al.
\newblock Learning transferable visual models from natural language supervision.
\newblock In \emph{International conference on machine learning}, pages 8748--8763. PmLR, 2021.

\bibitem[Schulman et~al.(2017)Schulman, Wolski, Dhariwal, Radford, and Klimov]{schulman_proximal_2017}
John Schulman, Filip Wolski, Prafulla Dhariwal, Alec Radford, and Oleg Klimov.
\newblock Proximal {Policy} {Optimization} {Algorithms}.
\newblock arXiv, August 2017.
\newblock \doi{10.48550/arXiv.1707.06347}.
\newblock arXiv:1707.06347 [cs].

\bibitem[Sharma et~al.(2021)Sharma, Azizan, and Pavone]{sharma2021sketching}
Apoorva Sharma, Navid Azizan, and Marco Pavone.
\newblock Sketching curvature for efficient out-of-distribution detection for deep neural networks.
\newblock In \emph{Uncertainty in artificial intelligence}, pages 1958--1967. PMLR, 2021.

\bibitem[Shi et~al.(2025)Shi, Ichter, Equi, Ke, Pertsch, Vuong, Tanner, Walling, Wang, Fusai, Li-Bell, Driess, Groom, Levine, and Finn]{shi_hi_2025}
Lucy~Xiaoyang Shi, Brian Ichter, Michael Equi, Liyiming Ke, Karl Pertsch, Quan Vuong, James Tanner, Anna Walling, Haohuan Wang, Niccolo Fusai, Adrian Li-Bell, Danny Driess, Lachy Groom, Sergey Levine, and Chelsea Finn.
\newblock Hi {Robot}: {Open}-{Ended} {Instruction} {Following} with {Hierarchical} {Vision}-{Language}-{Action} {Models}.
\newblock arXiv, February 2025.
\newblock \doi{10.48550/arXiv.2502.19417}.
\newblock arXiv:2502.19417 [cs].

\bibitem[Team et~al.(2025)Team, Abeyruwan, Ainslie, Alayrac, Arenas, Armstrong, Balakrishna, Baruch, Bauza, Blokzijl, et~al.]{team_gemini_2024}
Gemini~Robotics Team, Saminda Abeyruwan, Joshua Ainslie, Jean-Baptiste Alayrac, Montserrat~Gonzalez Arenas, Travis Armstrong, Ashwin Balakrishna, Robert Baruch, Maria Bauza, Michiel Blokzijl, et~al.
\newblock Gemini robotics: Bringing ai into the physical world.
\newblock \emph{arXiv preprint arXiv:2503.20020}, 2025.

\bibitem[Yang et~al.(2025)Yang, Garrett, Fox, Lozano-P\'erez, and Kaelbling]{yang2024guiding}
Zhutian Yang, Caelan Garrett, Dieter Fox, Tom\'as Lozano-P\'erez, and Leslie~Pack Kaelbling.
\newblock {Guiding Long-Horizon Task and Motion Planning with Vision Language Models}.
\newblock In \emph{ICRA}. IEEE, 2025.

\bibitem[Yuan et~al.(2024)Yuan, Duan, Blukis, Pumacay, Krishna, Murali, Mousavian, and Fox]{yuan_robopoint_2024}
Wentao Yuan, Jiafei Duan, Valts Blukis, Wilbert Pumacay, Ranjay Krishna, Adithyavairavan Murali, Arsalan Mousavian, and Dieter Fox.
\newblock {RoboPoint}: {A} {Vision}-{Language} {Model} for {Spatial} {Affordance} {Prediction} for {Robotics}.
\newblock arXiv, June 2024.
\newblock \doi{10.48550/arXiv.2406.10721}.
\newblock arXiv:2406.10721 [cs].

\bibitem[Zhao et~al.(2023)Zhao, Kumar, Levine, and Finn]{zhao_learning_2023}
Tony~Z. Zhao, Vikash Kumar, Sergey Levine, and Chelsea Finn.
\newblock Learning {Fine}-{Grained} {Bimanual} {Manipulation} with {Low}-{Cost} {Hardware}.
\newblock arXiv, April 2023.
\newblock \doi{10.48550/arXiv.2304.13705}.
\newblock arXiv:2304.13705 [cs].

\end{thebibliography}

\newpage
\appendix
\subsection{Low-Level Tracking Policy}\label{app:low-level}

In this section, we describe additional details on the low-level RL tracking policy.

Recall that the output of the low-level tracking policy consists of target joint positions that are fed into a PD controller. Given gains $k_p^i, k_d^i$, the torques $\tau_i$ on each joint $i$ are computed as:
$$\tau_i \;=\;
       k_p^i\,\Bigl(\,a_t^i \;-\; q_t^i\Bigr)
       \;\;-\;\;
       k_d^i\,\dot{q}_t^i.$$
The PD controller gains for each joint are shown in \cref{tab:pd_gains}

\begin{table}[ht]
\centering
\caption{PD controller gains for each joint.}
\label{tab:pd_gains}
\begin{tabular}{@{}lcc@{}}
\toprule
\textbf{Joint} & \boldmath{$k_p$} (Nm/rad) & \boldmath{$k_d$} (Nms/rad) \\
\midrule
\midrule
Hip Yaw        & 100 & 2.5 \\
Hip Roll       & 100 & 2.5 \\
Hip Pitch      & 100 & 2.5 \\
Knee           & 200 & 5.0 \\
Ankle Pitch    & 20  & 0.2 \\
Ankle Roll     & 20  & 0.1 \\
Shoulder Pitch & 90  & 2.0 \\

Shoulder Roll  & 60  & 1.0 \\
Shoulder Yaw   & 20  & 0.4 \\
Elbow          & 60  & 1.0 \\
Waist          & 400 & 5.0 \\
\bottomrule
\end{tabular}
\end{table}

\subsubsection{Policy design and training procedure}

The policy follows an actor-critic paradigm with a three-layer perception (MLP) backbone. Each hidden layer contains 256 units and ReLU activations. The system is trained by PPO in Isaac gym using a 0.02 integration timestep and 6144 parallel environments. The PPO hyperparameters used in training are shown in \cref{tab:ppo_hparams}.

\begin{table}[ht]
\centering
\caption{PPO hyperparameters used in training.}
\label{tab:ppo_hparams}
\begin{tabular}{@{}lc@{}}
\toprule
\textbf{Hyperparameter} & \textbf{Value} \\
\midrule
\midrule
Discount Factor ($\gamma$)             & 0.99 \\
GAE Parameter ($\lambda$)              & 0.95 \\
Timesteps per Rollout                  & 21 \\
Epochs per Rollout                     & 5 \\
Minibatches per Epoch                  & 4 \\
Entropy Bonus ($\alpha_{2}$)          & 0.01 \\
Value Loss Coefficient ($\alpha_{1}$) & 1.0 \\
Clip Range                             & 0.2 \\
Reward Normalization                   & Yes \\
Learning Rate                          & $1 \times 10^{-3}$ \\
\# Environments                        & 4096 \\
Optimizer                              & Adam \\
\bottomrule
\end{tabular}
\end{table}
To achieve robustness under real-world conditions, we randomize diverse physical parameters in training. Specifically, in each simulated episode, we include:
\begin{itemize}
    \item \textbf{Gravity Variation:} We sample gravity vectors near $(0,\,0,\,-9.81)$, shifting each component by up to \(\pm 0.1\). This accounts for inertial or slight calibration errors.
    \item \textbf{Friction Coefficients:} Each environment has friction coefficients drawn from [0.6,\,2.0], modeling differences in ground contact or footwear slip.
    \item \textbf{Base Mass and Center of Mass:} We add up to 5kg of virtual mass to the robot’s torso and shift its center of mass by up to 7cm. This simulates payloads or hardware variations.
    \item \textbf{Push Disturbances:} At random intervals, a lateral velocity impulse (up to 0.3m/s) disturbs the robot, prompting the policy to recover balance.
    \item \textbf{Motor Strength Range:} Each joint’s torque capacity is scaled by a factor in [0.8,\,1.2]. This reflects actuator performance variability.
    \item \textbf{Terrain Complexity:} Instead of a flat plane, we use a bumpy \emph{trimesh} terrain. The robot must learn to adapt foot placement and maintain stability on uneven ground.
\end{itemize}

At each time step, the instantaneous reward is computed as a weighted sum of individual terms. Specifically, this is a combination of tracking terms and regularization terms. These are shown in \cref{tab:reward_track} and \cref{tab:reward1}.

\begin{table}[h]
\centering
\caption{Final tracking reward terms and weights.}

\label{tab:reward_track}
\begin{tabular}{@{}lll@{}}
\toprule
\textbf{Term} & \textbf{Expression} & \textbf{Weight} \\
\midrule
\midrule
\multicolumn{3}{@{}l}{\textit{Expression Goal} $\mathcal{G}^{e}$} \\
DoF Position      & $\exp\bigl(-0.7\,\|\mathbf{q}_{\mathrm{ref}} - \mathbf{q}\|\bigr)$ & 3.0 \\
Keypoint Position & $\exp\bigl(-\|\mathbf{p}_{\mathrm{ref}} - \mathbf{p}\|\bigr)$      & 2.0 \\
\midrule
\multicolumn{3}{@{}l}{\textit{Root Movement Goal} $\mathcal{G}^{m}$} \\
Linear Velocity    & $\exp\bigl(-4.0\,\|\mathbf{v}_{\mathrm{ref}} - \mathbf{v}\|\bigr)$     & 6.0 \\
Velocity Direction & $\exp\bigl(-4.0\,\cos(\mathbf{v}_{\mathrm{ref}}, \mathbf{v})\bigr)$    & 6.0 \\
Roll \& Pitch      & $\exp\bigl(-|\mathbf{\Omega}_{\mathrm{ref}}^{\phi\theta} - \mathbf{\Omega}^{\phi\theta}|\bigr)$ & 1.0 \\
Yaw                & $\exp\bigl(-|\Delta y|\bigr)$                                          & 1.0 \\
\bottomrule

\end{tabular}
\end{table}

\begin{table}[h]
\centering
\caption{Regularization reward terms and weights.}
\label{tab:reward1}
\begin{tabular}{@{}lll@{}}
\toprule
\textbf{Term} & \textbf{Expression} & \textbf{Weight} \\
\midrule
\midrule
\multicolumn{3}{@{}l}{\textit{Feet Related}} \\
Height                 & $\max\bigl(|h_{\text{feet}}| - 0.2,\,0\bigr)$ & 2.0 \\
Feet Air Time          & $T_{\mathrm{air}}$                             & 10 \\
Drag                   & $\|\mathbf{v}_{\text{foot}}\| \sim \mathbf{1}_{\mathrm{new\_contact}}$ & $-0.1$ \\
Feet Contact Force     & $\|\mathbf{F}_{\mathrm{feet}}^{z}\| \ge 5\times \|\mathbf{F}_{\mathrm{feet}}^{x,y}\|$ & $-3 \times 10^{-3}$ \\
Stumble                & $\mathbf{1}\bigl\{\|\mathbf{F}_{\mathrm{feet}}\| > 4\times F_{\mathrm{feet}}^{z} \bigr\}$ & $-2$ \\
\midrule
\multicolumn{3}{@{}l}{\textit{Other Items}} \\
DoF Acceleration       & $\|\dot{\mathbf{q}}_{t}\|^{2}$                                     & $-3 \times 10^{-7}$ \\
Action Rate            & $\|\mathbf{a}_{t} - \mathbf{a}_{t-1}\|^{2}$                        & $-0.1$ \\
Energy                 & $\|\dot{\mathbf{q}}\|^{2}$                                         & $-1 \times 10^{-3}$ \\
Collision              & $\mathbf{1}_{\mathrm{collision}}$                                  & $-1$ \\
DoF Limit Violation    & $\mathbf{1}\{\,q_{t} \notin [q_{\min},\,q_{\max}]\}$               & $-10$ \\
DoF Deviation          & $\|\mathbf{q}_{t} - \mathbf{q}_{\mathrm{default}}\|^{2}$           & $-1.0$ \\
Vertical Lin. Velocity & $v_{z}^{2}$                                                        & $-1.0$ \\
Horiz. Ang. Velocity   & $\|\boldsymbol{\omega}_{xy}\|^{2}$                                 & $-2.0$ \\
Projected Gravity      & $\|\mathbf{g}_{xy}\|^{2}$                                          & $-2.0$ \\
\bottomrule
\end{tabular}
\end{table}

\subsubsection{Deployment evaluation}

\begin{figure}[ht] 
    \centering
    \includegraphics[width=\linewidth, trim={315px 0px 0px 0px}, clip]{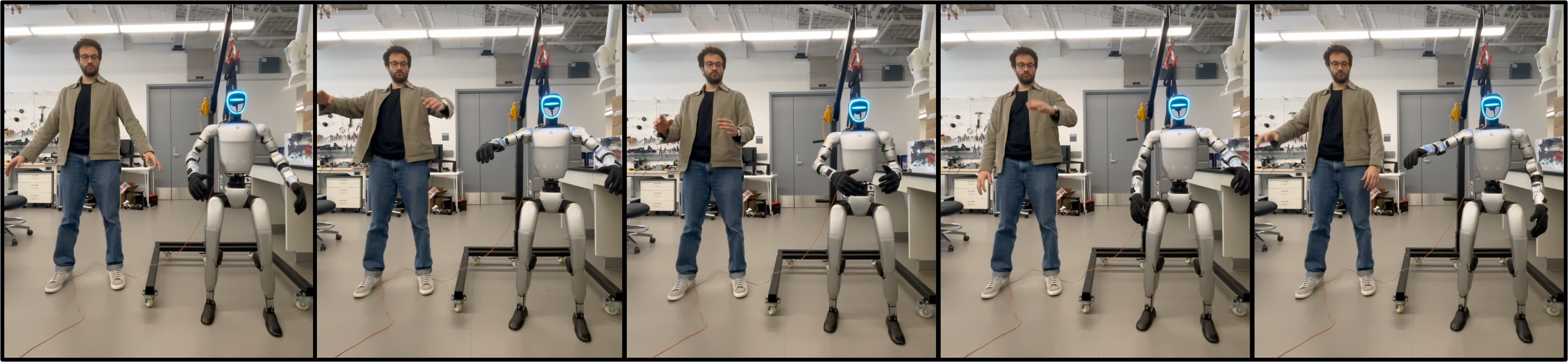}
    \caption{Real-time teleoperation on the physical Unitree G1 robot. A Human operator is providing real-time motion commands captured via an RGB camera, and the robot is executing the corresponding commands accurately, demonstrating robust and responsive real-world tracking performance.}
    \label{fig:real_robot_teleoperation}
\end{figure}

Following successful validation of the tracking policy in simulation, we deploy the policy on the physical Unitree G1 robot. Example real-time teleoperation is shown in \cref{fig:real_robot_teleoperation}. 

Quantitatively, we evaluate the tracking performance of our trained policy in simulation by computing the mean absolute error (MAE) for joint angles and keypoint positions over representative test motions. Specifically, we obtain an average joint-angle error of \(0.0593\,\mathrm{rad}\) (3.4\textdegree) and an average keypoint position error of \(13.88\,\mathrm{cm}\). These metrics indicate good tracking capabilities, validating our trained policy as suitable and reliable for real-world deployment on the humanoid robot.

\subsection{Mid-Level Imitation Learning}\label{app:mid-level}

In this section, we outline further implementation details for the mid-level imitation learning skill policies. 

\subsubsection{Teleoperation Architecture}

\begin{figure}[ht!]
    \centering
    \includegraphics[width=\linewidth, trim={0px 30px 0px 35px}, clip]{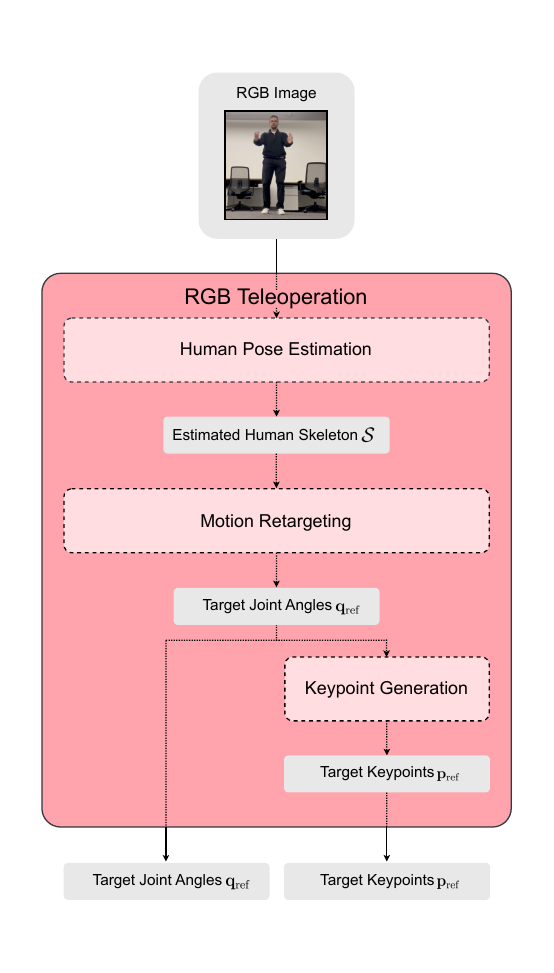}
    \caption{Overview of the Teleoperation Module pipeline. Human poses captured by an RGB camera are first estimated and then retargeted to the robot's kinematic structure. The resulting robot joint configurations are finally used to generate specific 3D keypoints required by the low-level tracking policy.}
    \label{fig:teleoperation_pipeline}
\end{figure}

An overall diagram for the RGB teleoperation architecture is shown in \cref{fig:teleoperation_pipeline}.

\begin{figure}[h]
    \centering
    \begin{subfigure}[b]{0.48\linewidth}
        \centering
        \includegraphics[width=0.8\linewidth, trim={100px 500px 100px 200px}, clip]{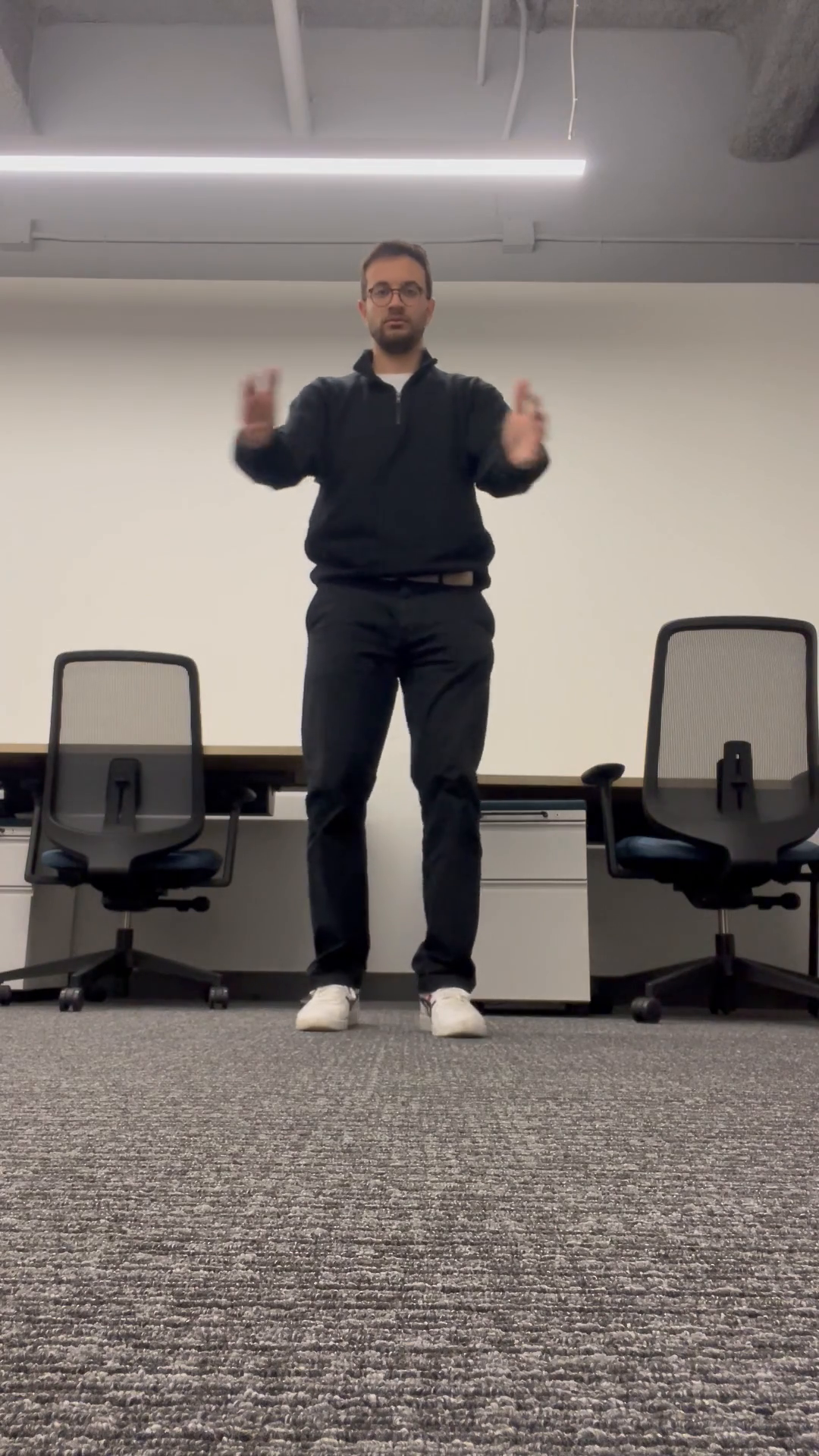}
        \caption{Human operator}
        \label{fig:hybrik1}
    \end{subfigure}
    \hfill
    \begin{subfigure}[b]{0.48\linewidth}
        \centering
        \includegraphics[width=0.8\linewidth, trim={100px 500px 100px 200px}, clip]{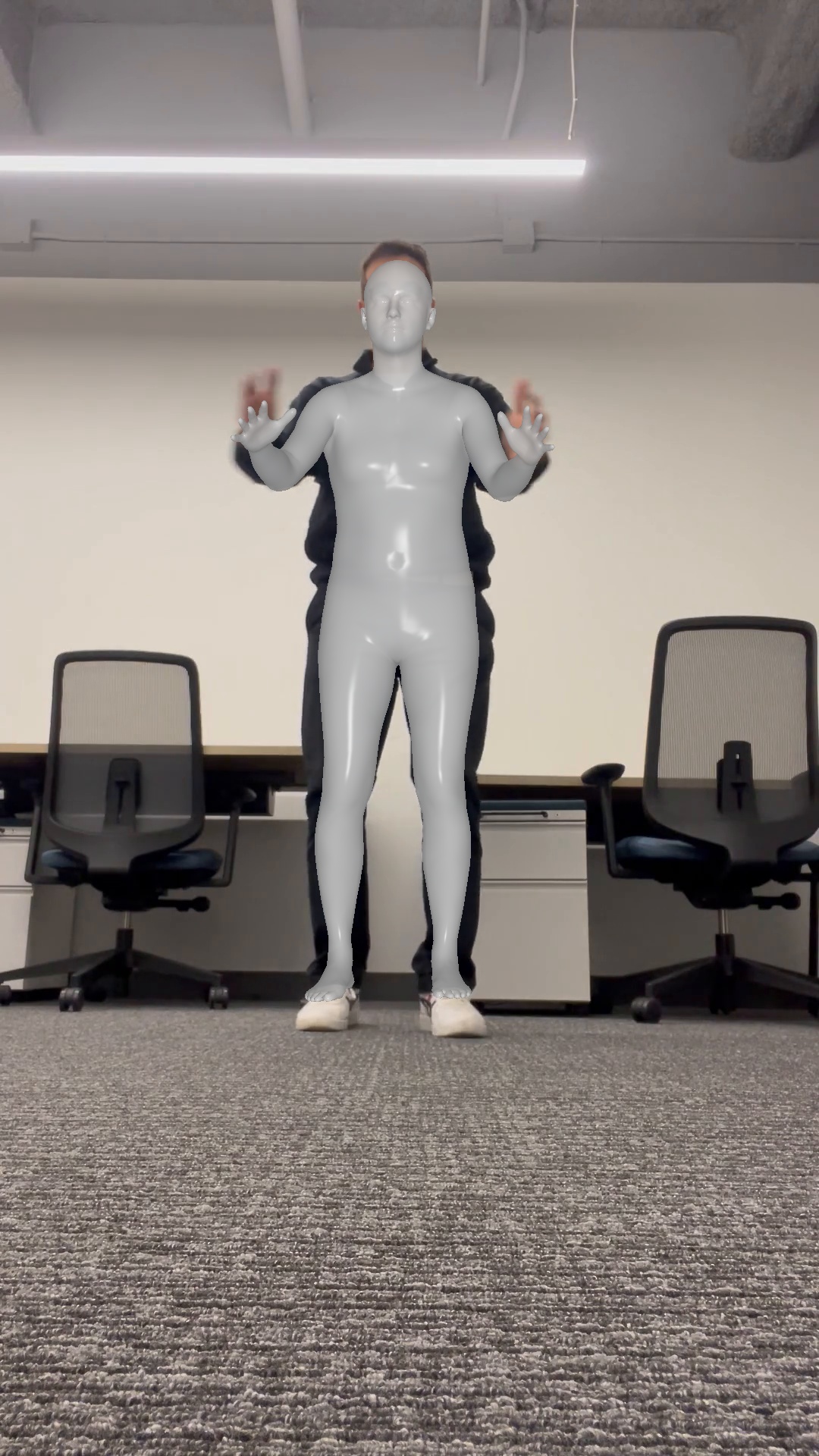}
        \caption{HybrIK pose detection result}
        \label{fig:hybrik2}
    \end{subfigure}
    \caption{Example of human pose detection by HybrIK, illustrating the original operator image (a) and corresponding detected skeletal configuration (b).}
    \label{fig:hybrik_detection}
\end{figure}

The joint retargeting procedure is outlined in \cref{alg:retargeting}  while an example of the human pose estimation used for teleoperation is shown in \cref{fig:hybrik_detection}.

\begin{algorithm}[h]
\caption{Retargeting procedure: human pose to robot configuration}
\label{alg:retargeting}
\begin{algorithmic}[1]
    \Statex \textbf{Input:} Source skeleton tree \(S\), source T-pose \(S_0\), target skeleton tree \(T\), joint mapping \(M\)
    \Statex \textbf{Output:} Retargeted skeleton state \(S_T\)

    \State \textbf{Joint filtering:}~~Remove joints from \(S\) and \(S_0\) that do not appear in the mapping \(M\);

    \State \textbf{Skeleton orientation alignment:}~~Compute alignment rotation \(R_{\text{align}}\) to align the human coordinate frame with the robot reference frame;
    \State Apply \(R_{\text{align}}\) to \(S\) and \(S_0\);

    \State \textbf{Root translation normalization:}~~Extract global root translation \(t\) from \(S\) and scale to match robot proportions;

    \State \textbf{Global rotation computation:}~~Compute relative rotation \(R_{\text{rel}}\) from the human T-pose \(S_0\) to the current pose \(S\);
    \State Apply \(R_{\text{rel}}\) to the robot T-pose to obtain \(R_{\text{global}}\);

    \State \textbf{Retargeted pose assembly:}~~Combine \(R_{\text{global}}\) and translation \(t\) to construct the retargeted pose \(S_T\);

    \State \textbf{Ground plane adjustment:}~~Adjust the z-axis of the translation in \(S_T\) to ensure feet contact the ground;

    \Statex \Return Retargeted skeleton state \(S_T\)
\end{algorithmic}
\end{algorithm}

\subsubsection{Imitation Learning Policy Details}

Hyperparameters for the Humanoid Imitation Transformer (HIT) \cite{fu_humanplus_2024} implementation are outlined in \cref{tab:hit_training_params}.

\begin{table}[ht]
    \centering
    \caption{Training hyperparameters for HIT model.}
    \label{tab:hit_training_params}
    \begin{tabular}{@{}lc@{}}
        \toprule
        \textbf{Hyperparameter} & \textbf{Value} \\
        \midrule
        \midrule
        Chunk size               & 50 \\
        Hidden dimension         & 512 \\
        Feedforward dimension    & 512 \\
        Transformer decoder layers & 6 \\
        Batch size               & 34 \\
        Learning rate            & \(1 \times 10^{-5}\) \\
        Training steps           & 100{,}000 \\
        Backbone network         & ResNet-18 (pretrained weights) \\
        Positional embeddings    & Enabled \\
        Feature loss weight      & 0.005 \\
        Encoder module           & Disabled (decoder-only architecture) \\
        Random seed              & 0 \\
        \bottomrule
    \end{tabular}
\end{table}

\subsection{Walk-through Example of our System Pipeline} \label{app:example}

\textbf{Example Pick-and-Place Task}. Consider the illustrative task of \emph{relocating an object from location A to location B} with the absence of obstacles. The hierarchical module operates as follows:

\begin{enumerate}
    \item The VLM Planner \(\Pi_{\mathrm{vlm}}^{\mathrm{plan}}\) generates a sequence of skills  \(\{\pi_1, \pi_2, \dots, \pi_k\}\) based on visual inputs and task prompts. In this example, the sequence would be \texttt{pick}, followed by \texttt{place}.
    \item The robot begins executing the first skill in the sequence, here the \texttt{pick} skill, governed by the trained imitation learning policy \(\pi_{\mathrm{pick}}\).
    \item Concurrently, the VLM Skill Monitor actively observes the robot's real-time visual feedback to assess whether the current skill (i.e., picking the object) has been completed. In this example, the monitor verifies whether the object is correctly grasped and above the target location.
    \item If the Skill Monitor indicates the skill remains \textit{in-progress}, the robot continues executing the current \texttt{pick} skill. Once the monitor confidently determines the completion of this skill, it signals the transition to the next planned skill.
    \item Upon receiving a \textit{completed} indication from the Monitor, the robot initiates the subsequent skill in the planned sequence—in this case, the \texttt{place} skill, governed by the corresponding policy \(\pi_{\mathrm{place}}\).
\end{enumerate}

\subsection{Example VLM prompts}\label{app:prompts}

Fig.~\ref{fig:system_prompt} shows the system prompt used in OpenAI API, Fig.~\ref{fig:task_prompt} shows the prompt for the planning task, while Fig.~\ref{fig:answer} shows an example GPT-4o response.

\begin{figure}[t]
    \centering
    \begin{lstlisting}
You are a helpful planning assistant for a robot. Your task is to create a sequence of actions (a plan) to achieve a given goal, based on the current environment shown in the image(s) and a set of available skills.

To achieve this, you will analyze the provided context, ground the skills, and MOST IMPORTANTLY, formulate the task of detecting skill success as a video question answering task. Each skill may have example questions to help ground the skills to the specific objects and actions in the context.

Output the plan as a JSON list, where each item in the list is an object representing a single step (a grounded skill). Each step object must have the following keys:
- "skill_name": The name of the skill used (from the available skills list).
- "description": A concise natural language description of *this specific action* being taken (e.g., "pick up the red block from the table").
- "preconditions": A natural language description of the state *required* just before executing *this specific action*, grounded to the objects involved (e.g., "the robot hand is empty and the red block is clear on the table").
- "effects": A natural language description of the state *resulting* from executing *this specific action*, grounded to the objects involved (e.g., "the robot hand is holding the red block").
- "question": A natural language question that can be used to verify if the action was successful - TRY TO FOLLOW SIMILAR QUESTION FORMATTING AND LANGUAGE AS THE EXAMPLE QUESTIONS PROVIDED IN THE SKILLS DESCRIPTION - the question should be specific and long enough to capture the essence of the action and its effects! 

If one of the preconditions or effects is IMPORTANT, it must be formulated as part of the question, so that the question can be used to verify the success of the action.

Carefully consider the preconditions, effects and example question wording of the available skills to ensure the plan is logical and feasible, and the question makes sense given the context.
    \end{lstlisting}
    \caption{Example system prompt for VLM skill planning.}
    \label{fig:system_prompt}
\end{figure}
    
\begin{figure}[t]
    \centering
    \begin{lstlisting}
Task:
    Pick up the bag and place it down on the white table.

Available Skills:
  - name: "pick"
    description: "Pick up an object."
    preconditions:
      - "hand is empty"
      - "object is on a surface"
    effects:
      - "hand is holding object"
      - "object is no longer on its previous surface"
      - "(IMPORTANTLY) object is held up as far to the left as possible"
    
  - name: "place"
    description: "Place a held object onto a surface."
    preconditions:
      - "hand is holding object"
      - "target surface is clear"
      - "target surface is reachable"
    effects:
      - "hand is empty"
      - "object is on target surface"
      
  - name: "push"
    description: "Push an object from one location to another."
    preconditions:
      - "hand is empty"
      - "object is pushable"
      - "object is reachable"
      - "target location is clear"
      - "object is at its start location"
    effects:
      - "object is at target location"
    
Generate the plan as a JSON list:
    \end{lstlisting}
    \caption{Example generated prompt for VLM skill planning.}
    \label{fig:task_prompt}
\end{figure}

\begin{figure}[t]
    \centering
    \begin{lstlisting}
[
SkillDescription(
    name='pick', 
    description='Pick up the bag from the box.', 
    preconditions="The robot's hand is empty and the bag is on the box.", 
    effects="The robot's hand is holding the bag up as far to the left as possible, and the bag is no longer on the box.", question='Has the robot finished picking up the bag and is holding the bag up to the left as far as possible?'
), 
SkillDescription(
    name='place', 
    description='Place the bag onto the white table.', 
    preconditions="The robot's hand is holding the bag, the white table is clear, and the table is reachable.", 
    effects="The robot's hand is empty and the bag is on the white table.", 
    question="Is the bag now placed on the white table and the robot's hand empty?"
)
]
    \end{lstlisting}
    \caption{Example VLM skill planning results.}
    \label{fig:answer}
\end{figure}

\end{document}